\documentclass[journal]{IEEEtran}
\usepackage[T1]{fontenc}
\usepackage{amsmath}
\usepackage{newtxtext,newtxmath}
\usepackage{graphicx}
\usepackage{booktabs}
\usepackage{multirow}
\usepackage[table]{xcolor}
\usepackage{cite}
\usepackage{url}
\usepackage[hidelinks]{hyperref}
\usepackage{pdfpages}

\newcommand{\BA}{\mathrm{BA}}
\newcommand{\dcross}{\Delta_{\mathrm{cross}}}
\newcommand{\ucross}{U_{\mathrm{cross}}}
\newcommand{\dsame}{\Delta_{\mathrm{same}}}
\newcommand{\pp}{\,pp}

\begin{document}
\title{Do Personal Adapters Carry Over Across Sessions? Separating Personal From Population Gains When Calibrating EEG Foundation Models for Motor Imagery}

\author{Xilin~Tao and Kani~Chen%
\thanks{X. Tao is with the Department of Mathematics, The Hong Kong University of Science and Technology, Hong Kong SAR, China (e-mail: xilin.tao@connect.ust.hk; ORCID: 0009-0001-1957-8406).}%
\thanks{K. Chen is with the Department of Mathematics and the Department of Industrial Engineering and Decision Analytics, The Hong Kong University of Science and Technology, Hong Kong SAR, China (corresponding author, e-mail: makchen@ust.hk).}%
\thanks{This work was supported in part by a Postgraduate Studentship from The Hong Kong University of Science and Technology (X. Tao).}}

\markboth{IEEE Transactions on Neural Systems and Rehabilitation Engineering}%
{Tao and Chen: Do Personal Adapters Carry Over Across Sessions?}

\maketitle

\begin{abstract}
Every new user of a motor imagery brain-computer interface must be calibrated, and users typically return on different days. A common approach with electroencephalography (EEG) foundation models fits a small personal adapter on a frozen backbone, but evaluations are usually confined to one session and rarely separate the user's own contribution from that of the population model. We quantified personal adapters with two paired contrasts, the net gain of a user's own adapter over the population model and its advantage over adapters fitted to other users, for three frozen foundation models (CBraMod, REVE and LaBraM). In 235 held-out subjects from three datasets, same-session calibration produced a positive net gain in all nine model and dataset combinations (means of 1.5 to 5.4 percentage points), and the gain remained positive when population training was quadrupled. In 54 two-session subjects, adapters fitted on the first day still improved on the population model on the second day by 1.31 points for REVE and 1.64 points for LaBraM (Holm-adjusted $p<0.01$), retaining 61\% and 83\% of the gain from same-day recalibration, whereas the gain for CBraMod was 0.04 points and indistinguishable from zero. Same-day recalibration gave the highest mean accuracy for all three models, and adding 5 to 20 second-day labels to the first-day adapter mostly fell short of it. Whether personal adapters carry over across sessions therefore depends on the model and must be tested per model, and personalization should be reported against both a population reference and exchanged parameters, including across sessions.
\end{abstract}


\section{Introduction}
\IEEEPARstart{M}{otor} imagery brain-computer interfaces (BCIs) translate the changes in EEG rhythms that accompany imagined limb movements into control commands, and they are an important route to rehabilitation training and assistive interaction. EEG differs markedly between people and between sessions of the same person, so a decoder usually requires each user to record a labeled calibration block before use \cite{krauledat2008}. Calibration takes time away from use and demands sustained attention, which is particularly costly for patients. Reducing or removing calibration has therefore been a central problem in the field \cite{gao2023,rao2026,wei2026}.

EEG foundation models pretrained on large collections of unlabeled recordings now provide reusable representations for downstream decoding \cite{wang2025cbramod,ouahidi2025reve,jiang2024labram,kim2024dfformer}. A common way to use them is to freeze the backbone, train a population model on many training subjects, and then fit a small set of personal parameters for each new user, for example a low-rank adaptation (LoRA) or a feature-wise modulation (FiLM) \cite{hu2022lora,perez2018film}. Gains from this strategy have been reported for motor imagery \cite{sarhane2026stacked,klein2025adapters,sirca2026plug,lopes2026published}. Recent calibration studies in this journal follow a related logic, either removing calibration with a large population model or adapting the population model with online self-training or a few labels from the same day \cite{gao2023,rao2026,leebh2024}.

However, these evaluations leave two questions open. First, the accuracy after adaptation mixes the contribution of the population model with that of the user's own data. If the reference model has a weaker head or less population training, personal parameters can appear beneficial even when they carry no information about the user, so paired comparisons against both the population model and other users' adapters are needed. Second, calibration and testing have almost always been performed within one session. In practice users come back on different days, and whether personal parameters fitted on one day remain useful on the next has not been tested directly for EEG foundation models. Cross-session distribution shift is well documented \cite{pang2024,chen2024abat}, but this alone does not reveal how much of a personal adapter captures stable properties of the user and how much fits the recording conditions of that day.

We address both questions with three frozen EEG foundation models. We first establish that same-session personal gains exist and test whether a stronger population model absorbs them. We then use two-session recordings to measure the net gain and the specificity of a first-day adapter on the second day, and compare it with second-day recalibration and with recalibration from a few labels. Our contributions are as follows.
\begin{enumerate}
\item We define a cross-session evaluation protocol for personalization that uses the population model and exchanged adapters as two paired references, specifies four contrasts (cross-session net gain, cross-session specificity, between-day loss and few-label recovery), and fixes a decision criterion in advance.
\item In 235 held-out subjects, three models and three datasets, we show that same-session personal gains are positive in all nine combinations and are not absorbed by a population model trained four times longer.
\item In 54 two-session subjects, we find that carry-over depends on the model. First-day adapters retain most of their gain and meet the decision criterion for REVE and LaBraM but give no net gain for CBraMod. Same-day recalibration is never worse on average, and a few second-day labels do not replace it.
\item We report absolute balanced accuracies for the three models and for two classical within-subject decoders, and release the code and subject-level results.
\end{enumerate}

\section{Related Work}
\subsection{EEG Foundation Models and Parameter-Efficient Adaptation}
EEG foundation models are pretrained with self-supervision on large recordings from many sources and montages, with the aim of learning representations that transfer across downstream tasks. CBraMod models spatial and temporal dependencies with criss-cross attention \cite{wang2025cbramod}, LaBraM segments signals into channel patches and learns a discrete neural codebook \cite{jiang2024labram}, and REVE uses a four-dimensional positional encoding to accept arbitrary electrode layouts and was pretrained on data from about 25,000 subjects \cite{ouahidi2025reve}. DFformer seeks pretraining that is not tied to a fixed data configuration \cite{kim2024dfformer}. Full fine-tuning of such models is expensive and prone to overfitting small calibration sets, so parameter-efficient adaptation has become the default. Subject-specific LoRA branches, stacked LoRA and subject-specific encoders have been applied to motor imagery decoding \cite{klein2025adapters,sarhane2026stacked,lopes2026published}, and adaptation strategies have been compared for plug-and-play use \cite{sirca2026plug}. These studies supply the adapter families and baselines used here, but they report overall accuracy after adaptation, with calibration and testing in the same session.

\subsection{Transfer Methods That Reduce Calibration}
Two lines of work reduce the calibration burden. The first strengthens the population model so that calibration can be skipped. Wei \textit{et al.} showed with 100 subjects that cross-subject accuracy increases with the number of training subjects \cite{wei2026}, Sartipi and Cetin built a subject-independent architecture that uses no target labels \cite{sartipi2024}, and Euclidean alignment reduces between-subject differences by whitening each subject's covariance \cite{he2020ea,wu2025revisit}. The second adapts the population model with a small amount of same-day data. Gao \textit{et al.} compared online self-training with few-label fine-tuning for a P300 speller \cite{gao2023}, Rao \textit{et al.} combined a subject-independent model with online supervised self-training to make a wearable motor imagery system calibration-free \cite{rao2026}, and Lee \textit{et al.} proposed an iteratively calibratable network \cite{leebh2024}. Across sessions, Pang \textit{et al.} combined multi-scale masked autoencoders with a few labels from the user for emotion recognition \cite{pang2024}, Chen \textit{et al.} evaluated alignment and adversarial training in block-wise and cross-session settings \cite{chen2024abat}, and Qin \textit{et al.} addressed distribution shift across devices \cite{qin2025}. This work establishes that cross-session shift is real and that same-day adaptation helps, but it does not test separately whether personal parameters remain useful on a later day.

\subsection{Attributing Personalization Gains}
Foundation model evaluation also faces a problem of attribution. Yang \textit{et al.} found that foundation models do not always outperform traditional decoders under the same protocol \cite{yang2026worth}, and audits have pointed out that overlap between pretraining corpora and downstream data blurs the meaning of an unseen subject \cite{lin2026identity,liu2026compass}. For personalization the problem is more specific, because the gain of a personal adapter may come from information about the user or from extra capacity and additional population training. We therefore do not propose a new adapter architecture. Instead, we build a paired evaluation on established LoRA and FiLM adapters that measures net gain against the population model and specificity against exchanged adapters, and we extend the same contrasts to the cross-session setting.

\section{Problem Formulation}
Let target subject $s$ be excluded from population training and validation, with one recording on each of two days $d\in\{1,2\}$. The labeled calibration trials of day $d$ are $\mathcal{C}_s^{(d)}$, and the scoring trials of day 2 are $\mathcal{Q}_s$, which follow $\mathcal{C}_s^{(2)}$ in time and are never used for fitting or selection. The frozen backbone is $f_\theta$ and the population parameters are $\phi$, comprising a shared adapter and a classification head that are learned on training subjects and selected on validation subjects. A personal adapter $a_s^{(d)}$ is fitted on $\mathcal{C}_s^{(d)}$ alone while $\phi$ is held fixed.

\begin{figure*}[!t]
\centering
\includegraphics[width=0.92\textwidth]{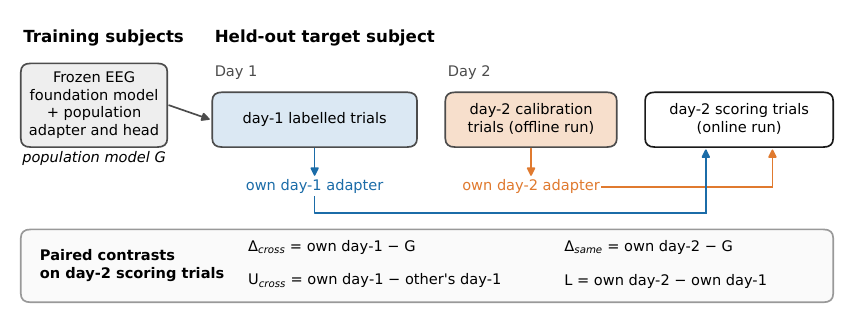}
\caption{Evaluation protocol. The population model $G$ is learned on training subjects and fixed together with the frozen backbone. The target subject's personal adapters are fitted on calibration trials from day 1 and day 2, and all conditions are compared on the day-2 online scoring trials. The four paired contrasts are the cross-session net gain $\dcross$, the cross-session specificity $\ucross$, the same-day gain $\dsame$ and the between-day loss $L$. In the same-session evaluation, the earlier and later halves of one recording take these roles and the contrasts are denoted $\Delta$ and $U$.}
\label{fig:protocol}
\end{figure*}

Let $\BA(\cdot)$ denote balanced accuracy on $\mathcal{Q}_s$, the mean of the per-class recalls, and let $G$ denote prediction with the population parameters only. For each target subject we define four contrasts. The cross-session net gain
\begin{equation}
\dcross=\BA(\phi,a_s^{(1)})-\BA(\phi)
\end{equation}
measures how much the first-day adapter improves on the population model on day 2. The cross-session specificity
\begin{equation}
\ucross=\BA(\phi,a_s^{(1)})-\BA(\phi,a_r^{(1)}),
\end{equation}
where $r$ is another test subject in the same fold, measures the advantage of the user's own adapter over another user's adapter. The same-day gain is $\dsame=\BA(\phi,a_s^{(2)})-\BA(\phi)$, and the between-day loss is
\begin{equation}
L=\BA(\phi,a_s^{(2)})-\BA(\phi,a_s^{(1)})=\dsame-\dcross .
\end{equation}
Fig.~\ref{fig:protocol} shows how these contrasts map onto the roles of the data.

Net gain and specificity are linked by the identity
\begin{equation}
U=\Delta-\left[\BA(\phi,a_r)-\BA(\phi)\right].
\end{equation}
If another user's adapter harms prediction, $U$ can be positive while $\Delta$ is zero or negative. Reporting $U$ alone would mistake the harm of a wrong adapter for a personal benefit, and reporting $\Delta$ alone cannot separate personal information from generic additional fitting, so we report both. In the same-session setting, $\mathcal{C}$ and $\mathcal{Q}$ are the earlier and later halves of one recording, and the contrasts are written $\Delta$ and $U$.

We also consider few-label recalibration. Starting from $a_s^{(1)}$, fitting continues with only the $n$ chronologically earliest day-2 labels to give $a_s^{(1+n)}$. The recovery is $R_n=\BA(\phi,a_s^{(1+n)})-\BA(\phi,a_s^{(1)})$, and it is compared with $G{+}n$, a new adapter fitted from the population model with the same $n$ labels. We assume that training and test subjects do not overlap and that no hyperparameter of the population model or of personal fitting is selected with the target's scoring trials. Conclusions are limited to two-class motor imagery, offline evaluation and the adapter families tested.

\textit{Decision criterion.} A model's personal adapters are judged to retain personal value across sessions when one-sided Wilcoxon tests of $\dcross$ and $\ucross$ are both below 0.05 after Holm correction. This criterion was fixed in a time-stamped protocol document on 27 October 2026 at 03:51 UTC, before the first cross-session run started at 05:12 UTC, and the protocol document is included in the code release.

\section{Methods}
\subsection{Population Model}
For each frozen backbone, the population model trains a shared adapter and a classification head on the backbone representation with a cross-entropy objective, using data from training subjects only. Checkpoints are selected by mean cross-entropy on validation subjects, with the initial checkpoint included as a candidate. To test whether the population model is sufficiently trained, training continues from the initially selected number of updates to twice and four times that number with the same objective, learning rate and sampling, and personal fitting hyperparameters are reselected at each endpoint. In the cross-session experiments the population model uses only first-day data from training subjects, and checkpoints are selected by the same validation rule within a bounded continuation budget.

\subsection{Personal Adapters}
Personal adaptation adds a new set of parameters for each target subject while the backbone, shared adapter and head remain fixed. LoRA adds a low-rank increment to an attention weight, $W\mapsto W+PQ^{\top}$, with $P$ initialized to zero so that predictions at the start of fitting equal those of the population model \cite{hu2022lora}. FiLM adds channel-wise scale and bias offsets to intermediate features, also initialized to zero \cite{perez2018film}. Personal fitting uses full-batch gradient updates whose learning rate, weight decay and number of steps are selected on validation subjects and then fixed for all target subjects. Zero initialization gives a testable implementation constraint, namely that an unfitted personal adapter must not change any prediction.

\subsection{Exchange Control}
To estimate $U$, each target subject draws ten donors with replacement from the other test subjects of the same dataset and fold. Each donor's adapter, fitted on the donor's own calibration data, is applied to the target, and balanced accuracy is averaged over donors on the target's scoring trials. The donor list is determined by the run seed and is shared across all compared conditions.

\subsection{Few-Label Recalibration}
Few-label recalibration continues fitting the first-day adapter with the earliest $n$ day-2 labels, with $n\in\{5,10,20\}$. Because the labels are taken in chronological order, a small set may contain only one class, and we keep such cases as they occur. The control $G{+}n$ fits a new personal adapter from the population model with the same $n$ labels.

\subsection{Statistical Inference}
For each contrast, training seeds are first averaged within subject to give one paired difference per subject. Subjects are then resampled with replacement 20,000 times, and the 2.5th and 97.5th percentiles give the 95\% confidence interval (CI) of the mean. $\dcross$ and $\ucross$ form one test family, evaluated with one-sided paired Wilcoxon signed-rank tests and Holm correction, and $L$ is tested two-sided. The intervals are conditional on the selected fits and do not account for the dependence introduced by shared population models and reused donors. A non-significant difference is interpreted as no detected difference, not as equivalence.

\section{Experimental Setup}
\subsection{Datasets and Splits}
The same-session evaluation uses three single-session datasets of left versus right hand imagery, PhysioNet EEGMMIDB \cite{schalk2009dataset,schalk2004bci2000}, Dreyer2023 \cite{dreyer2023dataset} and Cho2017 \cite{cho2017dataset}. After excluding subjects with sampling rate or duration anomalies and those flagged by the original authors, 103, 80 and 52 subjects remain, 235 in total. For each subject, trials ordered by run and onset are split into an earlier calibration half and a later scoring half. The cross-session evaluation uses Lee2019 \cite{lee2019dataset}, in which 54 subjects each recorded one session on each of two different days. Each session contains an offline run and an online run of 100 left and right hand trials each. The first-day adapter is fitted on the day-1 offline run, the second-day adapter on the day-2 offline run, and all conditions are scored on the day-2 online run. BCI Competition IV 2a \cite{tangermann2012competition} has 9 subjects with two sessions of 144 left and right hand trials each. Because of its size, it serves only as a descriptive supplement for CBraMod, with the first-day adapter fitted on the first half of the training session and scoring on the second half of the evaluation session. All cohorts are split by subject into five folds with disjoint training, validation and test subjects (Table~\ref{tab:data}), and each fold is run with five training seeds. Only public datasets were used. Ethical approval and informed consent for the original recordings were the responsibility of the original investigators, and no new data were collected.

\begin{table}[!t]
\centering
\caption{Datasets and Evaluation Settings}
\label{tab:data}
\footnotesize
\setlength{\tabcolsep}{2.5pt}
\begin{tabular}{lcccl}
\toprule
Dataset & $N$ & Sessions & Channels & Calib. / scoring trials \\
\midrule
PhysioNet & 103 & 1 & 64 / 27 & 22 / 23 \\
Dreyer2023 & 80 & 1 & 27 / 27 & 80--120 / 80--120 \\
Cho2017 & 52 & 1 & 64 / 27 & 100--120 / 100--120 \\
Lee2019 & 54 & 2 & 62 / 26 & 100 per day / 100 \\
BNCI2014-001 & 9 & 2 & 22 / 19 & 72 / 72 \\
\bottomrule
\end{tabular}

\vspace{2pt}
\parbox{\columnwidth}{\scriptsize Channels are EEG input / readout channels of the CBraMod pipeline. The three single-session datasets form one core cohort split into five folds of 169--171 training, 18 validation and 46--48 test subjects. Lee2019 folds have 39--40 training, 4 validation and 10--11 test subjects. Lee2019 is evaluated with all three models and BNCI2014-001 with CBraMod only.}
\end{table}

\subsection{Models and Implementation}
All three backbones use public pretrained weights and remain frozen throughout. CBraMod uses resampling to 200 Hz, a 0.3 Hz high-pass filter and a mains notch filter, and REVE uses the same shared preprocessing. LaBraM follows its original configuration with a 0.1 to 75 Hz band-pass filter and a 50 Hz notch filter. The personal LoRA rank is 8 for all three models. The CBraMod population model is a contextual model that generates FiLM parameters or LoRA mixing weights from resting-state context, and personal adaptation adds zero-initialized FiLM offsets or LoRA increments to it. The REVE and LaBraM population models are a shared attention LoRA with a classification head. Differences between the three models therefore combine differences in backbone and in population pipeline. All personal fitting hyperparameters were selected on validation subjects of the same-session core cohort and reused without tuning on Lee2019. Hyperparameter grids, the selected settings and the record of every cross-session run are given in Supplementary Sections S1 and S2. Experiments ran on NVIDIA RTX 3090 GPUs, and a Lee2019 run (one fold and one seed) took 0.08 h for CBraMod, 0.66 h for REVE and 0.20 h for LaBraM on average.

The reported pretraining sources of REVE include Dreyer, Cho and Lee (OpenBMI), those of LaBraM include PhysioNet, and those of CBraMod include none of the motor imagery datasets used here. This comparison relies on published source lists without record-level deduplication \cite{lin2026identity,liu2026compass}, so an unseen subject here means one that took no part in downstream population training and validation.

\subsection{Comparisons}
To test whether same-session personal gains exist, we compare the own adapter with the population model ($\Delta$) and with other users' adapters ($U$) in 235 subjects. To test whether a stronger population model absorbs the personal gain, we compare the change in $\Delta$ with the change in $G$ between one and four times the initial population training. To test whether personal adapters carry over across days, we compare on Lee2019 the first-day adapter with the population model ($\dcross$), with other users' first-day adapters ($\ucross$) and with second-day recalibration ($L$). To test whether a few second-day labels can replace recalibration, we compare the recalibrated adapter with full second-day calibration and with $G{+}n$. To place the results on an absolute scale, we use MOABB \cite{chevallier2024moabb} to implement two classical within-subject decoders, common spatial patterns with linear discriminant analysis (CSP+LDA) and tangent-space features with logistic regression (TS+LR), trained and scored on exactly the same trials as the personal adapters.

\section{Results}
\subsection{Same-Session Personal Gains Are Positive in All Combinations}
\begin{figure*}[!t]
\centering
\includegraphics[width=\textwidth]{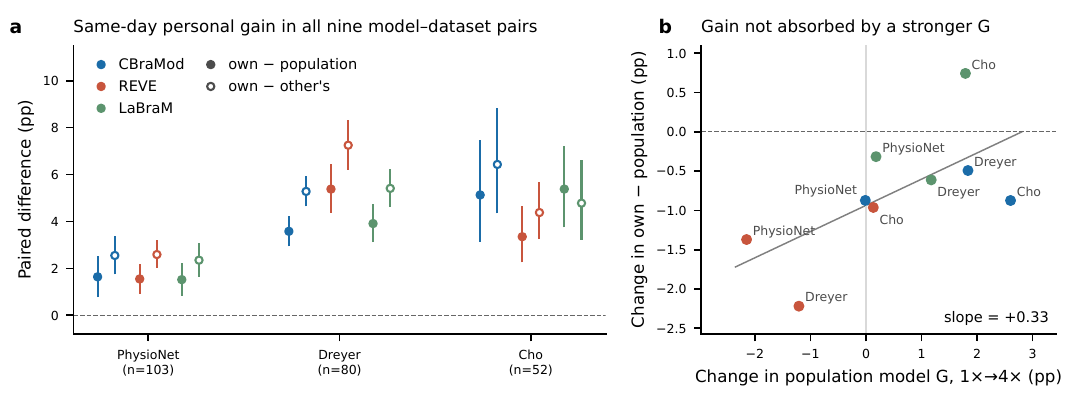}
\caption{Same-session personal gains. (a) Mean own minus population (filled) and own minus other users' adapters (open) for three models on three datasets. Error bars are subject-bootstrap 95\% CIs after averaging seeds within subject. (b) Change in the mean balanced accuracy of the population model $G$ against the change in own minus population when population training increases from one to four times the initial budget. Each point is one model and dataset combination, and the line is a least-squares fit. Panel (a) shows the values of Table~\ref{tab:same}. Panel (b) uses the population-budget extension, whose $1\times$ scores are computed on a separate scoring path with a separate bootstrap draw and differ from panel (a) by at most 0.04\pp.}
\label{fig:same}
\end{figure*}

In 235 held-out subjects, the own LoRA adapter outperformed both the population model and other users' adapters in all nine model and dataset combinations (Fig.~\ref{fig:same}a, Table~\ref{tab:same}). Mean net gains $\Delta$ ranged from 1.52 to 5.38 percentage points (pp) and mean specificities $U$ from 2.35 to 7.25\pp, with no 95\% CI including zero. The same pattern held in the six combinations whose dataset is not among the model's reported pretraining sources, namely CBraMod on all three datasets, REVE on PhysioNet and LaBraM on Dreyer and Cho. Only LaBraM on Cho had $U$ (4.78) smaller than $\Delta$ (5.38), which indicates that other users' adapters were on average slightly better than the population model there and that part of the personal fit transferred between people. Four further routes to a personal benefit were tested with CBraMod only, namely unlabeled rest context, nearest-donor selection, meta-learned initialization and Euclidean alignment. None gave a dependable improvement (Supplementary Section S4).

\begin{table*}[!t]
\centering
\caption{Same-Session Balanced Accuracy (\%) and Paired Contrasts (pp, 95\% CI) at the Initial Population Training Budget}
\label{tab:same}
\footnotesize
\begin{tabular}{llcccc}
\toprule
Model & Dataset & Population $G$ & Own adapter & $\Delta$ (own $-$ population) & $U$ (own $-$ other) \\
\midrule
\multirow{3}{*}{CBraMod} & PhysioNet & 80.99 & 82.63 & $+1.64$ [0.77, 2.51] & $+2.55$ [1.74, 3.40] \\
 & Dreyer & 80.37 & 83.95 & $+3.58$ [2.94, 4.24] & $+5.28$ [4.64, 5.93] \\
 & Cho & 64.16 & 69.29 & $+5.13$ [3.13, 7.46] & $+6.43$ [4.34, 8.83] \\
\midrule
\multirow{3}{*}{REVE} & PhysioNet & 84.71 & 86.26 & $+1.55$ [0.91, 2.19] & $+2.59$ [2.02, 3.19] \\
 & Dreyer & 83.91 & 89.29 & $+5.38$ [4.38, 6.43] & $+7.25$ [6.21, 8.33] \\
 & Cho & 69.77 & 73.12 & $+3.35$ [2.26, 4.65] & $+4.38$ [3.25, 5.70] \\
\midrule
\multirow{3}{*}{LaBraM} & PhysioNet & 82.29 & 83.82 & $+1.52$ [0.84, 2.24] & $+2.35$ [1.64, 3.08] \\
 & Dreyer & 83.17 & 87.08 & $+3.91$ [3.12, 4.76] & $+5.41$ [4.63, 6.25] \\
 & Cho & 60.70 & 66.08 & $+5.38$ [3.75, 7.20] & $+4.78$ [3.21, 6.62] \\
\midrule
CSP+LDA & PhysioNet / Dreyer / Cho & -- & 61.40 / 69.92 / 64.59 & -- & -- \\
TS+LR & PhysioNet / Dreyer / Cho & -- & 64.84 / 72.10 / 66.39 & -- & -- \\
\bottomrule
\end{tabular}

\vspace{2pt}
\parbox{\textwidth}{\scriptsize CSP+LDA and TS+LR are classical within-subject decoders trained on the same calibration trials and scored on the same trials. $n=103$, 80 and 52 subjects for PhysioNet, Dreyer and Cho.}
\end{table*}

\subsection{A Stronger Population Model Does Not Absorb the Gain}
If personal gains merely compensated for insufficient population training, they should shrink as the population model becomes stronger. When population training was quadrupled, $G$ improved by more than 1\pp\ in four of the nine combinations, by 2.60 and 1.84\pp\ for CBraMod on Cho and Dreyer and by 1.79 and 1.17\pp\ for LaBraM on Cho and Dreyer. In these four combinations $\Delta$ changed by $-0.87$ to $+0.74$\pp\ and remained positive (Fig.~\ref{fig:same}b). REVE showed the opposite case, with $G$ falling by 2.15 and 1.21\pp\ on PhysioNet and Dreyer and $\Delta$ falling with it. Across the nine combinations the slope of the change in $\Delta$ on the change in $G$ was $+0.33$, opposite in sign to absorption. At four times the initial budget, the pooled median gains of the three models remained between 1.0 and 2.0\pp, but no population model reached a confirmed plateau in training and validation loss, so this conclusion is limited to the tested range of training.

\subsection{Cross-Session Carry-Over Depends on the Model}
In 54 Lee2019 subjects, the first-day adapter outperformed other users' first-day adapters on day 2 for all three models (Table~\ref{tab:cross}), with $\ucross$ of $+2.92$ [1.53, 4.42]\pp\ for CBraMod, $+2.62$ [1.76, 3.51]\pp\ for REVE and $+2.75$ [1.82, 3.72]\pp\ for LaBraM (Holm-adjusted $p<0.001$ for all). The net gain diverged between models. For REVE and LaBraM, $\dcross$ was $+1.31$ [0.43, 2.23] and $+1.64$ [0.74, 2.59]\pp\ (Holm-adjusted $p=0.002$ and $p<0.001$), and 69\% and 70\% of subjects had a positive value, so both models met the decision criterion. For CBraMod, $\dcross$ was $+0.04$ [$-1.14$, 1.26]\pp\ ($p=0.540$) with 46\% of subjects positive, and the criterion was not met. The positive $\ucross$ of CBraMod arose because other users' adapters (67.13\%) fell below the population model (70.01\%), not because the own adapter exceeded it. The FiLM variant of CBraMod showed this more clearly, with $\dcross$ of $-1.81$ [$-3.42$, $-0.10$]\pp.

\begin{figure*}[!t]
\centering
\includegraphics[width=\textwidth]{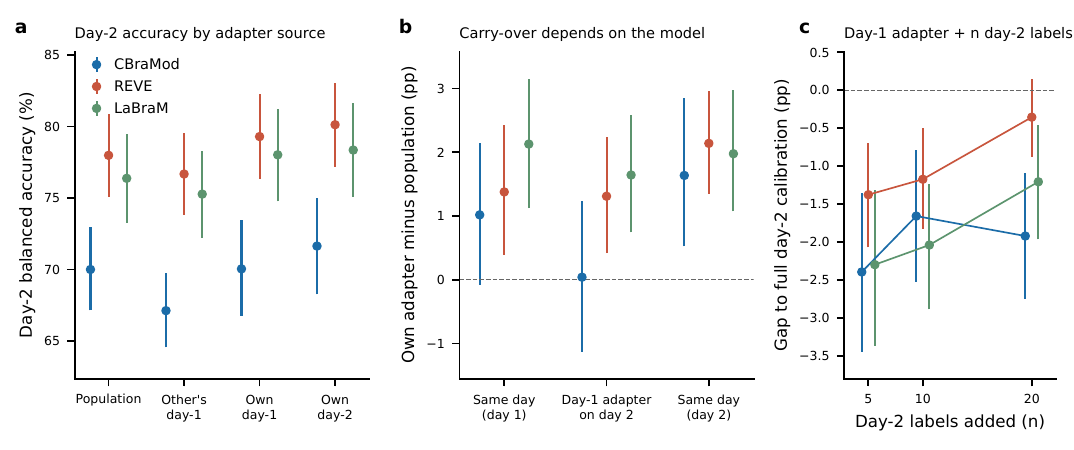}
\caption{Cross-session results on Lee2019 (LoRA, $n=54$). (a) Balanced accuracy on the day-2 online run for the population model, other users' day-1 adapters, the own day-1 adapter and the own day-2 adapter. (b) Own adapter minus population model, within day 1, for the day-1 adapter applied on day 2, and within day 2. (c) Gap to full day-2 calibration after adding $n$ day-2 labels to the day-1 adapter. Error bars are subject-bootstrap 95\% CIs.}
\label{fig:cross}
\end{figure*}

Relative to the same-day gain $\dsame$, the first-day adapter retained $1.31/2.14\approx61\%$ of the gain for REVE, $1.64/1.98\approx83\%$ for LaBraM and less than 3\% for CBraMod (Fig.~\ref{fig:cross}b). The between-day loss $L$ was $+1.60$ [0.61, 2.56]\pp\ for CBraMod ($p=0.002$), $+0.83$ [0.20, 1.47]\pp\ for REVE ($p=0.024$) and $+0.33$ [$-0.34$, 1.03]\pp\ for LaBraM ($p=0.521$). Second-day recalibration thus gave the highest mean accuracy for all three models, and the difference was significant for CBraMod and REVE. As a same-session reference, the gain of the first-day adapter on the day-1 online run, $\Delta_{\mathrm{S1}}$, was $+1.02$ [$-0.09$, 2.15]\pp\ for CBraMod, $+1.38$ [0.38, 2.44]\pp\ for REVE and $+2.13$ [1.13, 3.16]\pp\ for LaBraM. For CBraMod the first-day gain was therefore small even within day 1, with an interval including zero, and it vanished on day 2, whereas refitting on day-2 data still produced a gain.

BCI Competition IV 2a has only nine subjects and is reported descriptively. For CBraMod LoRA, $\dcross$ was $+2.10$ [$-0.52$, 4.91]\pp, $\ucross$ was $+5.49$ [2.95, 7.92]\pp\ in the eight subjects with an eligible donor (Holm-adjusted $p=0.017$), and $L$ was $+2.72$ [0.40, 5.22]\pp\ ($p=0.086$). The direction resembles that of REVE and LaBraM on Lee2019, with wider intervals.

\begin{table*}[!t]
\centering
\caption{Cross-Session Balanced Accuracy (\%) and Paired Contrasts (pp, 95\% CI)}
\label{tab:cross}
\scriptsize
\setlength{\tabcolsep}{3pt}
\begin{tabular}{lcccccccc}
\toprule
Model & $G$ & Own day 1 & Other day 1 & Own day 2 & $\dcross$ & $\ucross$ & $L$ & Criterion \\
\midrule
CBraMod & 70.01 & 70.05 & 67.13 & 71.65 & $+0.04$ [$-1.14$, 1.26] & $+2.92$ [1.53, 4.42] & $+1.60$ [0.61, 2.56] & Not met \\
REVE & 77.99 & 79.30 & 76.68 & 80.13 & $+1.31$ [0.43, 2.23] & $+2.62$ [1.76, 3.51] & $+0.83$ [0.20, 1.47] & Met \\
LaBraM & 76.39 & 78.03 & 75.28 & 78.36 & $+1.64$ [0.74, 2.59] & $+2.75$ [1.82, 3.72] & $+0.33$ [$-0.34$, 1.03] & Met \\
CSP+LDA & -- & 65.06 & -- & 69.15 & -- & -- & -- & -- \\
TS+LR & -- & 68.11 & -- & 70.94 & -- & -- & -- & -- \\
\midrule
CBraMod, BNCI$^{\dagger}$ & 64.17 & 66.27 & 61.35 & 68.98 & $+2.10$ [$-0.52$, 4.91] & $+5.49$ [2.95, 7.92] & $+2.72$ [0.40, 5.22] & -- \\
\bottomrule
\end{tabular}

\vspace{2pt}
\parbox{\textwidth}{\scriptsize $G$ is the population model. All rows except the last are the 54 Lee2019 subjects scored on the day-2 online run. $^{\dagger}$BCI Competition IV 2a, $n=9$, descriptive only ($\ucross$ uses the 8 subjects with an eligible donor). The criterion requires one-sided Wilcoxon tests of $\dcross$ and $\ucross$ to be below 0.05 after Holm correction. For the classical decoders, own day 1 and own day 2 are within-subject models trained on day-1 and day-2 calibration data.}
\end{table*}

\subsection{A Few Second-Day Labels Do Not Replace Recalibration}
Adding 5, 10 or 20 second-day labels to the first-day adapter gave no consistent improvement. The recovery $R_n$ was $-0.80$, $-0.06$ and $-0.33$\pp\ for CBraMod, $-1.97$, $-1.71$ and $-0.87$\pp\ for LaBraM and $-0.55$, $-0.34$ and $+0.47$\pp\ for REVE. All three intervals for LaBraM lay below zero, which indicates that a few new labels, possibly of a single class, can degrade an existing adapter. Compared with full second-day calibration, the gap after recalibration had an interval excluding zero in eight of the nine model and budget combinations (Fig.~\ref{fig:cross}c). The exception was REVE with 20 labels, with a gap of $-0.36$ [$-0.87$, 0.16]\pp. Fitting from the population model with the $n$ labels alone ($G{+}n$) gave a small gain for REVE ($+0.96$ [0.29, 1.61]\pp\ at $n=20$) and no detectable gain for CBraMod and LaBraM.

\subsection{Absolute Performance and Classical Decoders}
In the same-session setting, the population models of all three backbones exceeded both classical within-subject decoders by 8 to 23\pp\ on PhysioNet and Dreyer (Table~\ref{tab:same}). On Cho, the population models of CBraMod and LaBraM were below TS+LR, but with personal adapters all three models were comparable to or above it. In the Lee2019 cross-session setting, the population models of REVE and LaBraM (77.99\% and 76.39\%) exceeded TS+LR trained on same-day data (70.94\%), and that of CBraMod (70.01\%) was close to it (Table~\ref{tab:cross}). BCI Competition IV 2a was the only dataset on which a classical decoder exceeded the population model (TS+LR across sessions 70.83\%, CBraMod $G$ 64.17\%).

\section{Discussion}
\subsection{Why Carry-Over Differs Between Models}
For all three models the first-day adapter worked better for its owner than for others, so each adapter captured information about the user. The models differed in whether this information still produced a net gain over the population model on the next day. The first-day adapters of REVE and LaBraM retained most of their same-day gain with a small between-day loss, whereas that of CBraMod retained specificity without a net gain. An adapter can thus contain both a component that is stable for the user and a component specific to the recording conditions of the day, and the balance between them depends on the model.

Our design cannot identify the source of this difference, because at least three candidate factors are confounded with model identity. The first is the population pipeline. The CBraMod population model is conditioned on resting-state context, whereas REVE and LaBraM use a shared LoRA with a classification head, so personal adapters sit on different population parameters. The second is preprocessing, since the three models use different filter settings. The third is the level of the population model itself, which for CBraMod on Lee2019 was 6 to 8\pp\ below the other two models. Attributing the difference to backbone architecture would require repeating the experiment with a common population pipeline and preprocessing, which is the main direction for future work.

Pretraining overlap does not explain the difference. REVE's pretraining sources include Lee2019, which may raise its absolute accuracy, but LaBraM's sources do not, and LaBraM retained the largest share of its gain. Moreover, $\dcross$, $\ucross$ and $L$ are paired differences on the same frozen backbone, so pretraining effects on $G$ and on the personal adapter partly cancel.

\subsection{Implications for Calibration Practice}
The results have three practical implications. First, same-day recalibration was never worse on average than reusing an earlier adapter for any of the three models, so recalibrating in each session remains the safest choice when feasible. Second, reusing an earlier adapter to save calibration time must be validated for the model in use, because the same procedure worked for REVE and LaBraM but gave no net gain for CBraMod. Third, adding a few new labels to an old adapter does not replace full recalibration and degraded LaBraM. These findings agree with recent work showing that same-day online data benefit population models \cite{gao2023,rao2026}. Our addition is that the value of same-day data cannot simply be inherited from personal parameters fitted the day before.

For evaluation methodology, the gap between same-session and cross-session results shows that reporting personalization gains within a single session overstates their practical value. We recommend that personalization methods report both the net gain over the population model and the specificity against exchanged parameters, and report carry-over across days whenever two-session data are available. The CBraMod case shows that positive specificity alone does not imply a useful personal adapter.

\subsection{Limitations}
This study has several limitations. The three-model cross-session conclusions rest on 54 subjects from one two-session dataset, and BCI Competition IV 2a has nine subjects and was evaluated with CBraMod only. With two sessions, longer intervals and accumulation over many sessions were not examined. All evaluations were offline, used two-class motor imagery in healthy participants and did not include patient populations such as stroke survivors \cite{leem2024} or closed-loop feedback. Personal fitting hyperparameters were selected on same-session data and kept fixed, and cross-session tuning might change the results but would risk selection on the target dataset. The population models did not reach a confirmed plateau within the tested training range. Finally, as discussed above, the differences between models confound backbone, population pipeline and preprocessing.

\section{Conclusion}
We evaluated personal adapters on frozen EEG foundation models with two paired contrasts, against the population model and against exchanged parameters. Within a session, personal adapters produced positive net gains and specificity in all nine model and dataset combinations, and a stronger population model did not absorb the gain. Across sessions, first-day adapters retained most of their gain and met the pre-specified criterion for REVE and LaBraM but gave no net gain for CBraMod. Same-day recalibration was never worse on average than reusing an earlier adapter, and a few new labels did not replace full recalibration. Whether personal adapters carry over across days must be tested for each model, and evaluations of personalization should include the cross-session setting.

\section*{Data and Code Availability}
PhysioNet EEGMMIDB, Dreyer2023, Cho2017, Lee2019 and BCI Competition IV 2a are public and available from their original repositories or through MOABB. The analysis code of the same-session experiments is available at \url{https://github.com/gaivrt/eeg-personal-population-benefit} (version 1.0.0, DOI 10.5281/zenodo.22999685). An updated release containing the cross-session experiments, together with subject-level and seed-level results, is archived in the same repository (version 1.1.0, DOI 10.5281/zenodo.23807847).

\section*{Acknowledgment}
Computations were performed on the HPC4 cluster of The Hong Kong University of Science and Technology.

The authors used the AI system Claude (Anthropic) in this study. It was used to discuss the study design and analysis plan, to implement and run the cross-session experiments and statistical analyses on the authors' computing resources, to generate Figs.~1 to 3, to search for and check the cited literature, and to draft and revise the text of all sections. OpenAI Codex was used to implement the analysis code of the earlier same-session experiments. All reported numbers were produced by scripts from saved results. The authors reviewed the analyses, checked the reported numbers and references against the saved results and the cited sources, and take full responsibility for the content of this article.

\bibliographystyle{IEEEtran}
\bibliography{refs}

\clearpage
\includepdf[pages=-]{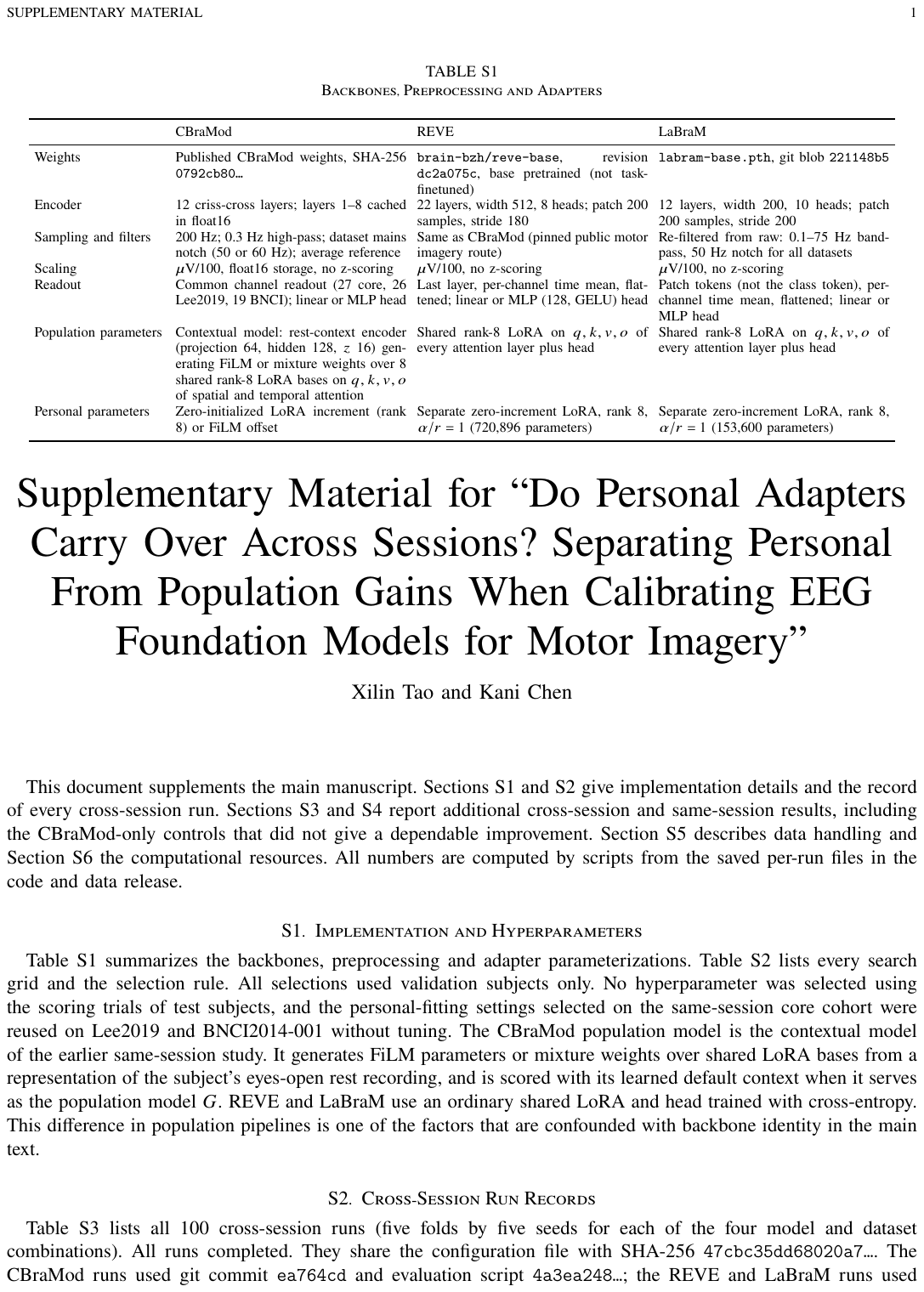}
\end{document}